\pdfoutput=1

\documentclass[letterpaper,10pt,conference]{ieeeconf}

\IEEEoverridecommandlockouts
\usepackage{graphicx}
\usepackage{amsmath}
\usepackage{amssymb}
\usepackage{newtxmath}
\usepackage{bm}
\usepackage{booktabs}
\usepackage{multirow}
\usepackage{array}
\usepackage{tabularx}
\usepackage{cite}
\usepackage[table]{xcolor}
\usepackage{url}
\usepackage{placeins}
\usepackage{balance}

\graphicspath{{images/}}

\makeatletter
\def\bstctlcite{\@bsphack\@bstctlcite}
\def\@bstctlcite#1{\@bsphack
  \@for\@citeb:=#1\do{%
    \edef\@citeb{\expandafter\@firstofone\@citeb}%
    \if@filesw\immediate\write\@auxout{\string\citation{\@citeb}}\fi}%
  \@esphack}
\makeatother

\newcommand{\vect}[1]{\bm{#1}}
\newcommand{\mat}[1]{\bm{#1}}
\definecolor{bestgray}{gray}{0.92}
\newcommand{\bestrow}{\rowcolor{bestgray}}
\newcolumntype{C}{>{\centering\arraybackslash}X}

\title{\LARGE \bf
OmniRisk: Omnidirectional Trajectory-Risk Learning for Agile Quadrotor Dynamic Avoidance
}

\author{Yifan He\textsuperscript{\textdagger}\textsuperscript{\rm 1},
Yang Liu\textsuperscript{\textdagger}\textsuperscript{\rm 1,2},
Wenhao Zhao\textsuperscript{\textdagger}\textsuperscript{\rm 1},
Hai Lin\textsuperscript{\rm 1},
Deping Zhang\textsuperscript{\rm 1},\\
Mingze Ma\textsuperscript{\rm 1},
Xin Zhou\textsuperscript{\rm 1},
Fei Gao\textsuperscript{\rm 1,2},
Huan Yu\textsuperscript{*}\textsuperscript{\rm 1,2},
Zipeng Dai\textsuperscript{*}\textsuperscript{\rm 1},
Ziming Ding\textsuperscript{*}\textsuperscript{\rm 1}\\[0.5ex]
{\normalsize \textsuperscript{\rm 1}Differential Robotics, Hangzhou, China \quad \textsuperscript{\rm 2}Zhejiang University, Hangzhou, China}\\
{\normalsize \textsuperscript{\textdagger}These authors contributed equally to this work. \quad \textsuperscript{*}Corresponding authors.}}

\begin{document}

\bstctlcite{IEEEexample:BSTcontrol}

\maketitle
\thispagestyle{empty}
\pagestyle{empty}

\begin{abstract}
Agile quadrotor avoidance of fast-moving obstacles requires anticipating collisions and selecting feasible maneuvers within short reaction windows.
Reliable predictive avoidance remains challenging because sparse range observations do not directly reveal obstacle motion, while online trajectory optimizers either scale poorly with obstacle count or remain efficient at the expense of reliability in dense, high-speed encounters.
We present OmniRisk, an omnidirectional planning framework that learns trajectory-level risk offline for efficient onboard evasion.
A fixed-dimensional tensor combines LiDAR range panoramas, dynamic masks, and Cartesian surface velocities to represent geometry and motion jointly.
We formulate an asymmetric risk field aligned with obstacle velocity that emphasizes approaching interactions and attenuates receding ones.
Accumulating this risk along predicted relative trajectories provides dense supervision and discourages unnecessary hesitation after obstacles pass.
A dual-branch circular convolutional network predicts terminal boundary states and dynamic risks for candidate primitives over an omnidirectional anchor lattice in a single forward pass, followed by selection and closed-form reconstruction of the selected candidate primitive.
This formulation removes online risk accumulation along trajectories and makes risk-inference cost independent of obstacle count.
OmniRisk enables efficient onboard avoidance, with real-world flights demonstrating consecutive evasive maneuvers at relative encounter speeds up to \(15\text{ m/s}\) without fine-tuning.
Code is available at \url{https://github.com/VANdexj/OmniRisk}.
\end{abstract}

\section{Introduction}
\label{sec:intro}

Navigating an agile quadrotor among high-speed dynamic obstacles is
fundamentally constrained by reaction latency.
The viable reaction window before impact is constrained by the platform's
kinodynamic limits and sensing-to-actuation delay.
Trajectory planning is the stage of this delay addressed in this work:
any time it consumes directly reduces the achievable evasive clearance.
Within this narrow reaction margin, the local planner must detect
omnidirectional threats and commit to dynamically feasible trajectories
before strict control deadlines expire.
Consequently, a planning pipeline whose latency scales with dynamic
obstacle count can incur severe execution lag or miss control
deadlines, precisely when evasive maneuvers are most critical.

\begin{figure}[!t]
    \centering
    \includegraphics[width=\columnwidth]{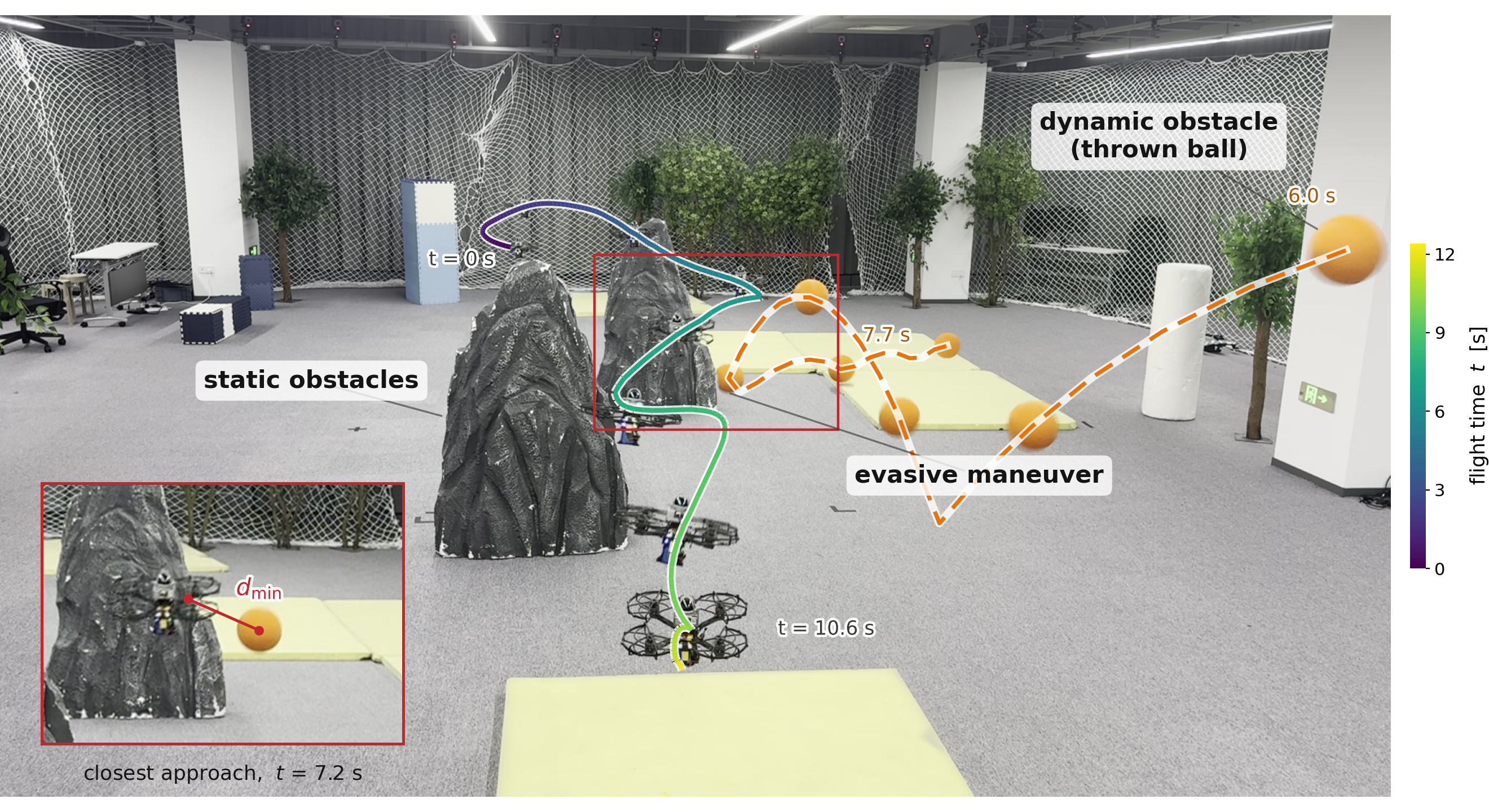}
    \caption{Real-world static--dynamic avoidance trial: a thrown ball crosses the
    quadrotor's route through rock-shaped static obstacles. The trajectory is
    colored by flight time; the inset marks the closest-approach distance
    $d_{\min}$.}
    \label{fig:real_world_teaser}
\end{figure}

Online numerical planners, which feed sparse range returns and predicted obstacle
trajectories directly into optimization or
search~\cite{tordesillas2022panther, huang2025safe, lu2024fapp},
face a structural trade-off between latency and safety.
Planners that enforce full spatiotemporal safety evaluate separating hyperplanes,
safe intervals, or collision constraints for every dynamic obstacle and time
sample~\cite{tordesillas2022panther, huang2025safe}, so their latency grows with
dynamic obstacle count $|\mathcal{O}_t|$ and overruns control deadlines.
Lightweight formulations~\cite{lu2024fapp} satisfy the timing budget by locally refining
a single trajectory against proximity penalties, but in our benchmark their
avoidance success drops sharply as threats become dense and fast.
Faster onboard compute shortens solve times but resolves neither limitation.

We resolve this trade-off by shifting continuous trajectory-risk
evaluation out of the onboard flight loop and into offline learning.
In simulation, each candidate primitive of an omnidirectional anchor lattice
is labeled with a static clearance cost and a dynamic risk computed from
ground-truth geometry and relative kinematics.
At runtime, the vehicle performs no online risk integration. Asynchronous LiDAR
returns and confirmed dynamic tracks are projected into a fixed-dimensional planning
input tensor. A dynamic head distilled from a relative-motion risk field then scores
every candidate primitive in a single forward pass, and the candidate primitive
with the lowest combined static and dynamic score is selected.
The corresponding trajectory is then recovered in closed form.
Iterative, obstacle-dependent optimization thus reduces to a single inference
whose cost is independent of $|\mathcal{O}_t|$, and because every candidate
primitive is scored, selection is not tied to one locally refined trajectory.

Fig.~\ref{fig:real_world_teaser} illustrates this in real-world flight, directly mapping physical elements
to underlying mechanisms: the vehicle traverses static rock clutter 
via the static clearance cost;
evades a crossing thrown ball at the marked closest approach $d_{\min}$ 
through the asymmetric dynamic risk field over Cartesian surface-velocity channels; 
and reaches the goal along a time-colored quintic trajectory 
without replanning stalls. 
A single forward pass jointly accounts for static clearance, 
dynamic evasion, and goal progress.

This paper makes three contributions:
\begin{itemize}
    \item A unified planning input tensor coupling a metric range panorama,
    a binary dynamic mask, and Cartesian surface-velocity channels, making
    obstacle motion directly observable under sparse LiDAR returns and enabling
    zero-shot deployment.
    \item An obstacle-velocity-aligned asymmetric dynamic risk field, continuous
    in the relative state, penalizing closing threats while attenuating
    the penalty on receding obstacles.
    \item A trajectory-risk learning paradigm replacing online numerical
    optimization with offline distillation, scoring all candidate primitives in
    a single forward pass with $\mathcal{O}(1)$ complexity in dynamic obstacle
    count $|\mathcal{O}_t|$ while retaining high avoidance success in dense,
    high-speed encounters.
\end{itemize}

\section{Related Work}
\label{sec:related}

\subsection{Omnidirectional Perception and the Planning Interfaces}
\label{sec:rw_representation}

Existing dynamic avoidance frameworks rely primarily on spatiotemporal corridors
or time-indexed occupancy grids~\cite{wang2021autonomous, kong2021avoiding, lu2022perception, zhong2024safer, huang2025safe},
yet the cost of constructing these structures grows with obstacle density,
so planning latency rises with scene complexity.
While omnidirectional LiDAR systems~\cite{wu2025flying} mitigate sensor blind spots,
passing raw occupancy grids and bounding-box lists downstream forces planners
to reconstruct collision geometry online.
Conversely, event-based avoidance methods~\cite{he2021fastdynamicvision, falanga2020dynamic, sanket2020evdodgenet}
achieve microsecond sensing latency
but generate unconstrained single-step reflexes
that lack metric scale and dynamically feasible planning horizons.
To resolve this trade-off, we consolidate omnidirectional metric range,
dynamic cluster masking, and Cartesian surface velocities
into a fixed-dimensional planning input tensor,
yielding a metric-scale input
whose size is independent of dynamic obstacle count.

\subsection{Spatiotemporal Risk Fields and Trajectory Optimization}
\label{sec:rw_risk}

Lightweight penalty formulations~\cite{lu2024fapp} remain fast but refine a
single trajectory locally; in our benchmark this limits success in dense,
high-speed encounters (Sec.~\ref{sec:exp_closed_loop}).
Formulations that enforce full spatiotemporal safety through per-obstacle constraints
or safe-interval search~\cite{tordesillas2022panther, huang2025safe}
instead incur costs that scale with obstacle count, risking deadline misses.
Analytical collision metrics such as closest point of approach (CPA)
and time to collision (TTC) attempt to alleviate this burden,
yet evaluating exact clearance along high-order polynomials lacks closed-form solutions,
forcing fragile online numerical root finding over crude spherical bounds.
Furthermore, while velocity-dependent potential fields and motion-shaped rewards~\cite{ginesi2021dynamic, xu2025flow}
modulate repulsive forces,
they remain causally symmetric---penalizing receding obstacles identically to closing threats---or
only adjust the current control step without anticipating future interactions.
Instead of fragile online optimization or symmetric pointwise potentials,
we formulate an obstacle-velocity-aligned asymmetric dynamic risk field
evaluated via discretized path integrals offline,
converting predicted relative motion
into continuous training labels for a feed-forward network.

\subsection{Learning-Based Low-Latency Avoidance}
\label{sec:rw_learning}

Learning-based planners circumvent iterative online optimization,
yet agile dynamic evasion simultaneously demands metric spatial reasoning,
kinodynamically feasible outputs,
and computational complexity decoupled from obstacle count.
Existing frameworks satisfy only subsets of these requirements.
Imitation learning pipelines~\cite{tordesillas2023deeppanther}
distill trajectories from computationally prohibitive numerical experts,
inheriting the shortcomings of expert demonstrations while enforcing kinodynamic limits
solely through soft penalties.
Alternative approaches rely on forward-facing depth~\cite{lu2024you}
or sector-partitioned range inputs~\cite{xu2025flying}
that lack explicit surface-velocity channels,
yielding unconstrained single-step reactive actions~\cite{fan2025flying, zhang2026threataware}
without explicit metric safety margins.
In contrast, our framework unifies static geometry and Cartesian surface velocities
within a metric planning input tensor,
evaluating candidate primitives over a fixed omnidirectional anchor lattice
in a single forward pass
to generate minimum-jerk primitives with kinodynamically bounded terminal states.

\section{Problem Formulation and System Overview}
\label{sec:framework}

\begin{figure*}[!t]
    \centering
    \includegraphics[width=\textwidth]{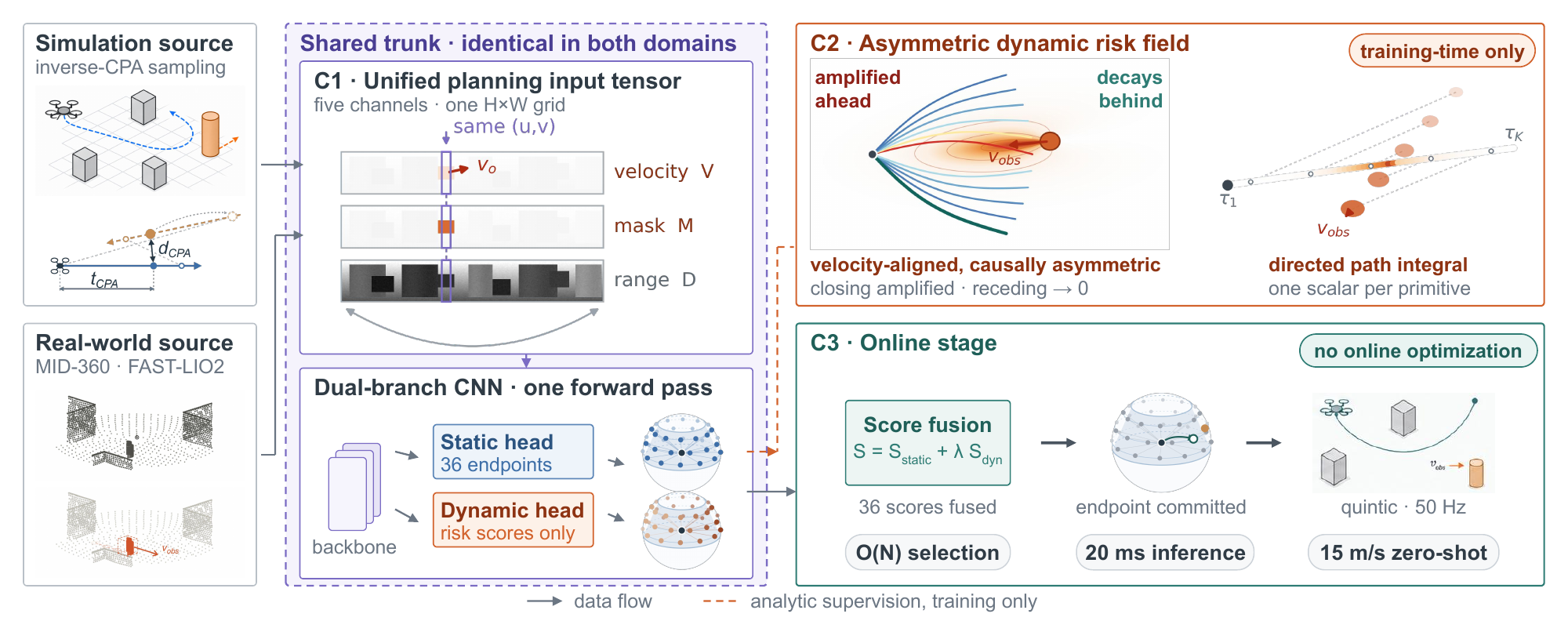}
    \caption{System architecture. Offline, the candidate primitives $\Pi$ are
    supervised by the static clearance cost and the asymmetric dynamic risk
    field (C2, dashed arrows). Online, streaming perception is rasterized into
    the unified planning input tensor $\mat{I}_t$ (C1); a single forward pass over
    the omnidirectional anchor lattice $\mathcal{A}$ scores all candidate primitives for
    $20\,\mathrm{Hz}$ trajectory commitment without online optimization (C3).}
    \label{fig:framework}
\end{figure*}

\subsection{Problem Formulation}
\label{sec:pf}

\subsubsection{Reformulation}
\label{sec:pf_recast}
High-speed evasion is governed by a strict timing budget:
to prevent execution lag and collision,
the local planner must commit a dynamically feasible trajectory
before the current replanning cycle ends.
Numerical planners struggle to reconcile this budget with safety:
full spatiotemporal constraints incur runtimes that grow with
dynamic obstacle count $|\mathcal{O}_t|$, whereas lightweight local refinement
remains fast but degrades in dense, high-speed encounters (Sec.~\ref{sec:intro}).
Rather than accelerating either form of iterative optimization,
we recast real-time avoidance as a feed-forward evaluation
over a pre-defined omnidirectional anchor lattice,
shifting spatiotemporal risk integration entirely into offline training.
A single forward pass then scores every candidate primitive:
trajectory-risk evaluation becomes independent of $|\mathcal{O}_t|$,
selection no longer depends on one locally refined trajectory,
and planning latency follows a fixed-size forward pass, not $|\mathcal{O}_t|$.

\subsubsection{Notation}
\label{sec:pf_notation}
Let $\mathcal{W}$ denote the inertial world frame and $\mathcal{P}_t$ 
the gravity-aligned local planning frame anchored at the vehicle at perception time $t$. 
The platform state is defined as $\vect{x}_0 = (\vect{p}_0, \vect{v}_0, \vect{a}_0)^\top \in \mathcal{X} \subset \mathbb{R}^9$, 
bounded by kinodynamic limits $\|\vect{v}_0\|_2 \le v_{\max}$ and $\|\vect{a}_0\|_2 \le a_{\max}$. 
Given a navigation goal $\vect{g}$, the perception front-end streams 
a non-repetitive LiDAR point set $\mathcal{L}_t$, an odometric state estimate $\vect{x}_t$, 
an accumulated static background, and a set of confirmed dynamic obstacle tracks
$\mathcal{O}_t = \{(\vect{c}_o, \vect{v}_o, \vect{s}_o, \iota_o)\}_{o=1}^{|\mathcal{O}_t|}$, where each track is parameterized by centroid $\vect{c}_o$,
metric velocity $\vect{v}_o$, bounding extent $\vect{s}_o = (d_{o,x}, d_{o,y}, d_{o,z}) \in \mathbb{R}^3$, 
and persistent tracking identity $\iota_o$.

\subsubsection{Planning Input Tensor}
Privileged simulation $\mathcal{D}_{\mathrm{sim}}$ provides ground-truth states
and future trajectory risk, whereas real deployment $\mathcal{D}_{\mathrm{real}}$ does not.
The policy therefore receives only a calibrated planning input tensor in both domains:
\begin{equation}
\mat{I}_{t} = [\mat{D}_{t}, \mat{M}_{t}, \mat{V}_{t}] \in \mathbb{R}^{H \times W \times 5},
\label{eq:planning_input_tensor}
\end{equation}
thereby strictly isolating all privileged simulation labels from deployment.
Over an $H \times W$ spherical grid,
$\mat{I}_{t}$ concatenates a metric range panorama $\mat{D}_{t} \in \mathbb{R}^{H \times W \times 1}$
recording nearest surface distances,
a binary dynamic mask $\mat{M}_{t} \in \{0,1\}^{H \times W \times 1}$
identifying returns assigned to dynamic tracks $\mathcal{O}_{t}$,
and Cartesian surface-velocity channels $\mat{V}_{t} \in \mathbb{R}^{H \times W \times 3}$
encoding metric obstacle velocities $\vect{v}_{o}$ on dynamic pixels and zeros elsewhere.
This fixed-dimensional tensor decouples representation complexity
from raw point-cloud density and dynamic obstacle count $|\mathcal{O}_{t}|$,
with cross-domain channel invariance detailed in Sec.~\ref{sec:domain_invariance}.

\subsubsection{Action Space}
\label{sec:pf_action}
The action space is built on an \emph{omnidirectional anchor lattice}
$\mathcal{A} = \{\hat{\vect{a}}_i\}_{i=1}^{N}$, a fixed set of $N = 36$
unit-direction anchors expressed in $\mathcal{P}_t$.
$\mathcal{A}$ comprises 12 azimuth bins that uniformly partition the full $360^\circ$
and three elevation bands that span the sensor field of view.
Omnidirectional here refers to full $360^\circ$ azimuthal coverage,
the property exploited by the circular convolutions of Sec.~\ref{sec:distill_arch}.
Neither the anchors nor their count $N$ varies at runtime.
Anchored to $\mathcal{A}$ are the \emph{candidate primitives}
$\Pi = \{\vect{\pi}_i\}_{i=1}^{N}$ defined over a planning horizon $T$.
Each candidate primitive $\vect{\pi}_i(t): [0, T] \to \mathbb{R}^3$
is parameterized as a minimum-jerk quintic polynomial uniquely determined
by the initial state $\vect{x}_0 = (\vect{p}_0, \vect{v}_0, \vect{a}_0)^\top$
and a terminal boundary state $\vect{x}_f^{(i)} = (\vect{p}_f^{(i)}, \vect{v}_f^{(i)}, \vect{a}_f^{(i)})^\top$
confined to the angular cell of anchor $\hat{\vect{a}}_i$.
$\mathcal{A}$ thus fixes the anchors and $N$,
while the terminal boundary state within each angular cell adapts to the observed static geometry
through the static head of Sec.~\ref{sec:distill_isolation}.
Candidate primitives violating the kinodynamic limits $\|\vect{v}(t)\|_2 \le v_{\max}$
or $\|\vect{a}(t)\|_2 \le a_{\max}$ are marked infeasible and excluded during selection,
while the network output remains fixed at $N$ candidates.

\subsubsection{Online Objective}
\label{sec:pf_objective}
Conditioned on the planning input tensor $\mat{I}_t$ and navigation goal $\vect{g}$, 
trajectory selection reduces to identifying the selected candidate primitive
$\vect{\pi}^* \in \Pi$ minimizing a composite objective:
\begin{equation}
\begin{aligned}
\vect{\pi}^* = \arg\min_{\vect{\pi}_i \in \Pi} \!\big[&J_{\mathrm{prog}}(\vect{\pi}_i, \vect{g})
+ \lambda_s J_{\mathrm{smooth}}(\vect{\pi}_i) \\
&+ J_{\mathrm{risk}}(\vect{\pi}_i, \mat{I}_t)\big],
\end{aligned}
\label{eq:selection_objective}
\end{equation}
where $J_{\mathrm{prog}}$ penalizes goal displacement
and $J_{\mathrm{smooth}}$ regulates control effort via integrated squared jerk;
both admit closed-form evaluation.
Crucially, $J_{\mathrm{risk}}$ is a learned surrogate evaluated via feed-forward inference.
It estimates the accumulated spatiotemporal risk of the asymmetric dynamic field
directly from $\mat{I}_t$, bypassing iterative online optimization.
\subsection{System Framework}
\label{sec:sf}

\subsubsection{Offline Stage}
\label{sec:sf_offline}
Fig.~\ref{fig:framework} summarizes the resulting two-stage architecture.
The offline phase constructs the supervision needed to bypass real-time
numerical optimization.
Within the privileged simulation domain $\mathcal{D}_{\mathrm{sim}}$, the training
pipeline samples randomized dynamic obstacle encounters and accumulates
collision risk along candidate primitives.
Specifically, inverse-CPA sampling first draws a target closest-approach distance and time (\(d_{\mathrm{CPA}}, t_{\mathrm{CPA}}\)) 
and back-computes the obstacle's initial state,
concentrating training samples in near-collision regimes.
A compact neural network is trained via supervised regression to predict
this continuous trajectory risk directly from the synthesized
planning input tensor $\mat{I}_t$, embedding non-convex collision reasoning into
the network weights prior to deployment and eliminating online numerical
integration.

\subsubsection{Online Stage}
\label{sec:sf_online}
The onboard framework comprises three asynchronously decoupled modules:
an omnidirectional perception front-end, the trained dual-branch circular convolutional network,
and an $\mathrm{SO}(3)$ geometric tracking controller. 
At each cycle, the front-end integrates streaming LiDAR points $\mathcal{L}_t$ 
and confirmed dynamic tracks $\mathcal{O}_t$ into $\mat{I}_t$ at $20\,\mathrm{Hz}$. 
Synchronized to this representation, the static and dynamic heads score
the candidate primitives $\Pi$ in parallel at $20\,\mathrm{Hz}$,
selecting $\vect{\pi}^*$
via~\eqref{eq:selection_objective}.
Concurrently, the $\mathrm{SO}(3)$ controller tracks $\vect{\pi}^*$ at $50\,\mathrm{Hz}$.
Owing to this asynchronous decoupling, a delayed planning cycle does not stall control;
the controller continues tracking the previously committed trajectory.

\section{Omnidirectional Perception Representation}
\label{sec:perception}

\subsection{Spherical Projection and the Metric Range Panorama}
\label{sec:spherical_projection}

\subsubsection{Range Panorama Construction}
\label{sec:sd_separation}
FAST-LIO2~\cite{xu2022fastlio2} provides
LiDAR--inertial odometry and motion compensation.
To structure unorganized point clouds while preserving full $360^\circ$ spatial coverage,
raw returns $\mathcal{L}_t$ are projected onto a sphere
centered at the origin of $\mathcal{P}_t$.
Each 3D point $\vect{p} = (x, y, z)^\top \in \mathcal{L}_t$ is mapped to spherical coordinates $(\theta, \phi, r)$, 
where azimuth $\theta = \mathrm{atan2}(y, x) \in [-\pi, \pi)$, 
elevation $\phi = \arcsin(z / \|\vect{p}\|_2) \in [-\phi_{\min}, \phi_{\max}]$, 
and Euclidean range $r = \|\vect{p}\|_2$. 
Discretizing azimuth and elevation into an $H \times W$ grid yields
the metric range panorama $\mat{D}_t \in \mathbb{R}^{H \times W \times 1}$,
with each cell $(u, v)$ recording the minimum measured range along that bearing:
\begin{equation}
\mat{D}_t(u, v) = \min_{\{\vect{p} \in \mathcal{L}_t \mid \lfloor \kappa(\vect{p}) \rfloor = (u, v)\}} \|\vect{p}\|_2,
\label{eq:range_projection}
\end{equation}
where $\kappa: \mathbb{R}^3 \to [0, H-1] \times [0, W-1]$ 
denotes the continuous spherical-to-pixel projection operator.

\subsubsection{Temporal Depth Completion}
\label{sec:depth_completion}
To mitigate the angular sparsity inherent in non-repetitive scanning LiDARs,
the front-end maintains a sliding temporal buffer of motion-compensated sweeps 
over $\Delta t_{\mathrm{acc}}$. Unobserved rays in $\mat{D}_t$ are completed 
via edge-preserving nearest-neighbor propagation capped at $r_{\max}$. 
This produces a dense, metrically consistent depth panorama capturing both 
thin obstacles and free-space boundaries with negligible computational overhead.

\subsection{Dynamic Masking and Velocity Rasterization}
\label{sec:tracking}

To maintain spatial and temporal consistency with offline training supervision, 
dynamic tracking estimates obstacle positions and velocities at the planning time 
within the local frame $\mathcal{P}_t$. 

M-detector~\cite{wu2024mdetector}, an event-based point-stream detector,
segments moving returns from the accumulated static background.
The resulting dynamic clusters are associated across successive frames
via the Hungarian algorithm based on Euclidean centroid distances, 
maintaining a six-dimensional state $(\vect{c}_o, \vect{v}_o) \in \mathbb{R}^6$ per track. 
A track hypothesis is confirmed after consistent temporal associations 
over a lifespan $\tau_{\text{track}}$, 
with extrapolation bounded within $\tau_{\text{pred}}$. 

Given measurement delivery latency $\Delta t = t_{\text{plan}} - t_{\text{meas}}$, 
a closed-form ballistic model propagates each measured state to the planning time:
\begin{equation}
  \vect{c}_o^r = \vect{c}_o^m + \vect{v}_o^m \Delta t + \frac{1}{2} \vect{g} \Delta t^2, \quad 
  \vect{v}_o^r = \vect{v}_o^m + \vect{g} \Delta t,
  \label{eq:ballistic_sync}
\end{equation}
where superscripts $m$ and $r$ denote raw and synchronized states, respectively, 
and $\vect{g}$ accounts for vertical gravitational acceleration during the latency interval. 

These synchronized states are subsequently mapped into 
the dynamic mask $\mat{M}_t$ and velocity channels $\mat{V}_t$ of $\mat{I}_t$.

\subsection{Domain Invariance and Sim-to-Real Transfer}
\label{sec:domain_invariance}

\subsubsection{Channel Alignment}
\label{sec:di_manifold}
To maintain a consistent interface between $\mathcal{D}_{\mathrm{sim}}$ and $\mathcal{D}_{\mathrm{real}}$,
the perception front-end keeps all five channels
numerically and geometrically consistent.
Both domains share identical spherical discretization ($H \times W$),
circular boundary conditions across the azimuthal seam,
and matched background accumulation windows.
Crucially, depth completion is restricted exclusively to the range channel $\mat{D}_{t}$,
preventing spatial dilation from corrupting
the binary dynamic mask $\mat{M}_{t}$ or Cartesian surface-velocity channels $\mat{V}_{t}$.
This consistency ensures that $\mat{I}_{t}$ encodes the same geometric quantities in both domains,
enabling zero-shot policy deployment.

\subsubsection{Mask-Channel Corruption}
\label{sec:di_corruption}
During simulation training, stochastic noise is introduced into the dynamic mask $\mat{M}_t$ 
via independent false-positive ($p_{\text{fp}}$) 
and false-negative ($p_{\text{fn}}$) perturbations. 
Spurious mask regions are assigned zero velocity in $\mat{V}_t$, 
whereas unmasked dynamic surfaces retain their metric velocities. 
This decoupling discourages the policy from relying solely on binary occupancy cues, 
guiding the network to evaluate trajectory risk 
primarily through Cartesian surface kinematics.

\section{Trajectory Planning and Risk Learning}
\label{sec:planning}

\subsection{Static Clearance Cost}
\label{sec:static_cost}

Within the privileged simulation domain $\mathcal{D}_{\mathrm{sim}}$, the static clearance cost is evaluated against an analytical Euclidean signed distance field (ESDF) in the local planning frame $\mathcal{P}_t$.
Each candidate primitive is sampled at $K$ uniform collocation points $\tau_j = \frac{j}{K}T$, yielding
\begin{equation}
  J_{c,i} = \frac{1}{K} \sum_{j=1}^K \exp \left( -\frac{d(\vect{\pi}_i(\tau_j)) - d_{\text{safe}}}{\sigma_{\text{static}}} \right),
  \label{eq:static_cost}
\end{equation}
where $d(\cdot)$ denotes the distance to the nearest static boundary, $d_{\text{safe}}$ is the target clearance, and $\sigma_{\text{static}}$ scales the decay.
This exponential penalty maintains non-vanishing repulsive gradients within nominally safe margins, biasing trajectory selection toward wider corridors.

\subsection{Asymmetric Dynamic Risk Field}
\label{sec:risk_field}

\begin{figure}[!tb]
    \centering
    \includegraphics[width=\columnwidth,trim=40.8bp 15.6bp 15.6bp 21bp,clip]{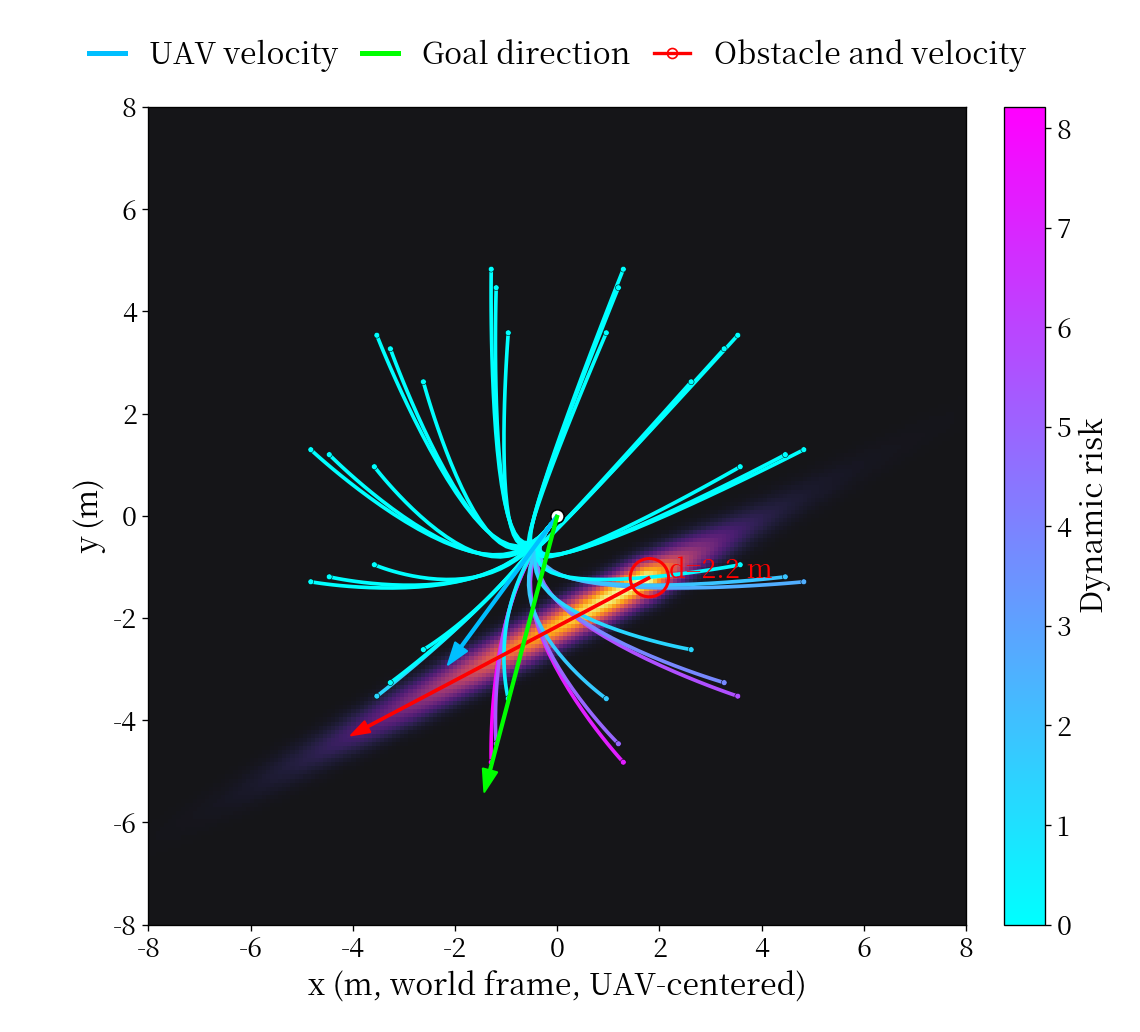}
    \caption{Asymmetric dynamic risk field in the horizontal plane. The
    quadrotor at the origin (velocity $\vect{v}_0$, blue) targets the goal (green),
    encountering an obstacle (velocity $\vect{v}_o$, red) at $d = 2.2$~m. Background:
    pointwise dynamic risk $r_o$ under relative obstacle motion.
    Candidate primitives are colored by dynamic risk
    $J_{d,i}$~\eqref{eq:primitive_dynamic_cost}.}
    \label{fig:risk}
\end{figure}

\subsubsection{Limitations of Proximity Metrics}
\label{sec:risk_precedents}
Dynamic avoidance formulations that evaluate collision risk
primarily through instantaneous spatial proximity or isotropic inflation~\cite{wang2021autonomous, falanga2020dynamic, lu2024fapp}
suffer from three fundamental kinematic limitations.
First, they are causally symmetric,
penalizing receding obstacles identically to closing threats,
so the penalty persists behind already-cleared obstacles.
Second, they are isotropic,
failing to elongate along-track safety margins proportionally to obstacle speed.
Third, rigid distance thresholds produce step-like collision labels
whose gradients vanish almost everywhere, destabilizing network regression.
These structural deficiencies motivate an obstacle-velocity-aligned,
causally asymmetric risk field that provides dense,
continuous supervision over relative obstacle motion.

\subsubsection{Anisotropic Relative Metric}
\label{sec:risk_anisotropy}
To evaluate interaction kinematics over the horizon, we express the relationship
between candidate primitive $\vect{\pi}_{i}(t)$ and obstacle $o$, with synchronized state
$(\vect{c}_{o}^{r}, \vect{v}_{o}^{r})$, in relative coordinates:
\begin{equation}
\vect{q}(t) = \vect{\pi}_{i}(t) - (\vect{c}_{o}^{r} + \vect{v}_{o}^{r}t), \quad
\vect{v}_{\mathrm{rel}}(t) = \dot{\vect{\pi}}_{i}(t) - \vect{v}_{o}^{r}.
\label{eq:relative_kinematics}
\end{equation}
A constant-velocity forecast enables closed-form evaluation,
and the spatial metric aligns with the travel axis
$\vect{e}_{o} = \vect{v}_{o}^{r} / \max(\|\vect{v}_{o}^{r}\|_{2}, \epsilon_{v})$,
where $\epsilon_{v} > 0$ prevents numerical singularity.
The anisotropic metric tensor is formulated as
\begin{equation}
\mat{A}_{o} = \sigma_{\perp}^{-2} \mat{I}_{3\times3} + 
(\sigma_{\parallel}^{-2} - \sigma_{\perp}^{-2}) \vect{e}_{o} \vect{e}_{o}^{\top},
\label{eq:anisotropic_metric}
\end{equation}
where lateral scale $\sigma_{\perp}$ reflects obstacle geometry 
and longitudinal scale $\sigma_{\parallel}$ expands with speed $\|\vect{v}_{o}^{r}\|_{2}$. 
This yields an anisotropic Gaussian base $\phi_{o}(\vect{q}) = \exp(-\vect{q}^{\top} \mat{A}_{o} \vect{q})$, 
elongating along-track clearance while preserving lateral bounds.

\subsubsection{Softplus Causal Gate}
\label{sec:risk_selector}
Although anisotropic, $\phi_o(\vect{q})$ is symmetric ($\phi_o(\vect{q})=\phi_o(-\vect{q})$):
at the same relative position it cannot distinguish approaching from receding obstacles;
the horizon sum~\eqref{eq:primitive_dynamic_cost} separates the two only through future distances.
To enforce causal asymmetry, relative velocity is projected onto
$\vect{n}_o = \vect{g}_o / \max(\|\vect{g}_o\|_2, \epsilon_g)$,
where $\vect{g}_o = \nabla_{\vect{q}} \phi_o = -2\phi_o \mat{A}_o \vect{q}$ is the inward normal of the iso-risk ellipsoid
and $\epsilon_g > 0$ again prevents numerical singularity.
With $z_o = \alpha \vect{v}_{\mathrm{rel}}^\top \vect{n}_o$ ($\alpha > 0$), the pointwise risk is
\begin{equation}
r_o(\vect{q}, \vect{v}_{\mathrm{rel}}) = \phi_o(\vect{q}) \operatorname{softplus}(z_o)
= \phi_o(\vect{q}) \ln(1 + e^{z_o}).
\label{eq:dynamic_risk_field}
\end{equation}
Since $\dot{\vect{q}} = \vect{v}_{\mathrm{rel}}$, $z_o > 0$ exactly when $\vect{q}^\top \mat{A}_o \vect{q}$ decreases,
which agrees with the sign of the Euclidean range rate if $\sigma_{\parallel} = \sigma_{\perp}$ or $\vect{q} \parallel \vect{e}_o$ or $\vect{q} \perp \vect{e}_o$.
Approaching risk is amplified and receding risk decays toward zero (Fig.~\ref{fig:risk}),
so a crossing obstacle is penalized mainly before its closest approach;
deceleration or re-approach violates the constant-velocity forecast and is reflected only after the tracked velocity is updated.

\subsubsection{Directed Path Integral}
\label{sec:risk_integral}
The dynamic risk of candidate primitive $\vect{\pi}_i$ aggregates the pointwise risk:
\begin{equation}
  J_{d,i} = \sum_{j=1}^K \sum_{o \in \mathcal{O}_t} r_o \left( \vect{q}(\tau_j),\, \vect{v}_{\mathrm{rel}}(\tau_j) \right).
  \label{eq:primitive_dynamic_cost}
\end{equation}
With uniform $\tau_j$, $J_{d,i}$ is a Riemann sum of the time-integrated risk scaled by $K/T$,
a constant absorbed into $w_{\mathrm{scale}}$.
Since the quadrotor and each obstacle are evaluated at the same instants $\tau_j$,
$J_{d,i}$ mainly penalizes candidate primitives on a collision course.

\subsection{Dual-Branch Network Distillation}
\label{sec:distillation}

\subsubsection{Network Architecture}
\label{sec:distill_arch}
A dual-branch circular convolutional network evaluates terminal boundary states
and dynamic risks in a single forward pass.
The static branch applies a convolutional backbone to the range panorama $\mat{D}_{t}$
to extract environmental geometry,
while the dynamic branch processes $[\mat{M}_{t}, \mat{V}_{t}]$
to capture the Cartesian motion field.
Convolutions are circularly padded along the azimuth dimension
to preserve $360^{\circ}$ continuity across coordinate seams.
Both heads also receive $(\vect{v}_{0}, \vect{a}_{0})$ and the goal $\vect{g}$ in $\mathcal{P}_{t}$,
with $\vect{g}$ rescaled to at most unit length.
The static head uses only the $\mat{D}_{t}$ feature to infer terminal boundary states $\vect{x}_{f}^{(i)}$
and static costs $\hat{J}_{s,i}$,
while the dynamic head adds the $[\mat{M}_{t}, \mat{V}_{t}]$ feature to predict dynamic risks $\hat{J}_{d,i}$.

\subsubsection{Head Isolation}
\label{sec:distill_isolation}
Candidate primitive geometry is generated exclusively by the static head;
the dynamic head only evaluates it.
Since $\vect{x}_{f}^{(i)}$ depends only on $\mat{D}_{t}$, $(\vect{v}_{0}, \vect{a}_{0})$, and $\vect{g}$
and its objective excludes dynamic risk,
$[\mat{M}_{t}, \mat{V}_{t}]$ cannot alter candidate primitive geometry;
the dynamic loss reaches the shared $\mat{D}_{t}$ encoder only through $\hat{J}_{d,i}$.
This isolation suppresses high-frequency trajectory chattering
and promotes spatial consistency across consecutive replanning cycles.

\subsubsection{Training Objective}
\label{sec:distill_loss}

Terminal boundary states are trained by directly minimizing smoothness,
static clearance cost~\eqref{eq:static_cost}, anchor alignment, and acceleration effort,
excluding dynamic risk.
The static head regresses the total static cost $J_{s,i}$,
which replaces anchor alignment with goal progress,
using a smooth-$L_{1}$ loss.
The dynamic head is trained in a compressed logarithmic domain:
\begin{equation}
\mathcal{L}_{\mathrm{risk}} = \frac{1}{N} \sum_{i=1}^{N} l_{\text{smooth-}L_{1}}\!\left(\hat{J}_{d,i},\, \mathrm{sg}\!\left[\ln(1 + w_{\mathrm{scale}} J_{d,i})\right]\right),
\end{equation}
where $w_{\mathrm{scale}}$ balances dynamic penalties.
Logarithmic scaling compresses orders-of-magnitude risk variations across diverse velocities,
preventing near-miss extremes from dominating gradients.
The three losses are summed with equal weights and optimized jointly without freezing either head.
Each minibatch recomputes $J_{s,i}$ and $J_{d,i}$ on the current candidate primitives
from the privileged ESDF and ground-truth tracks,
and the stop-gradient $\mathrm{sg}[\cdot]$ blocks gradients through both targets.

\subsection{Online Selection}
\label{sec:selection}

At runtime, the two heads evaluate all $N$ candidate primitives
in a single forward pass,
selecting the candidate index $i^*$ according to
\begin{equation}
i^* = \arg\min_{i} \left(\hat{J}_{s,i} + \lambda \hat{J}_{d,i} + \mathcal{R}_{\mathrm{con}}(i)\right),
\end{equation}
where $\lambda$ scales dynamic risk
and $\mathcal{R}_{\mathrm{con}}(i)$ enforces temporal continuity and switching hysteresis
to suppress trajectory chattering.
Selection is $\mathcal{O}(N)$ in $N$ but $\mathcal{O}(1)$ with
respect to dynamic obstacle count because $N=36$ is fixed.
Here, $\hat{J}_{s,i}$ absorbs $J_{\mathrm{prog}}$, $J_{\mathrm{smooth}}$, and static clearance cost,
while $\lambda\hat{J}_{d,i}$ instantiates $J_{\mathrm{risk}}$ in~\eqref{eq:selection_objective}.
The polynomial coefficients of $\vect{\pi}^* = \vect{\pi}_{i^*}$ are then computed in closed form
and mapped to attitude-thrust setpoints for tracking.

\section{Experimental Validation}
\label{sec:experiments}

We systematically evaluate the proposed framework across simulated
and real-world environments.

\begin{figure}[!t]
    \centering
    \includegraphics[width=0.85\columnwidth]{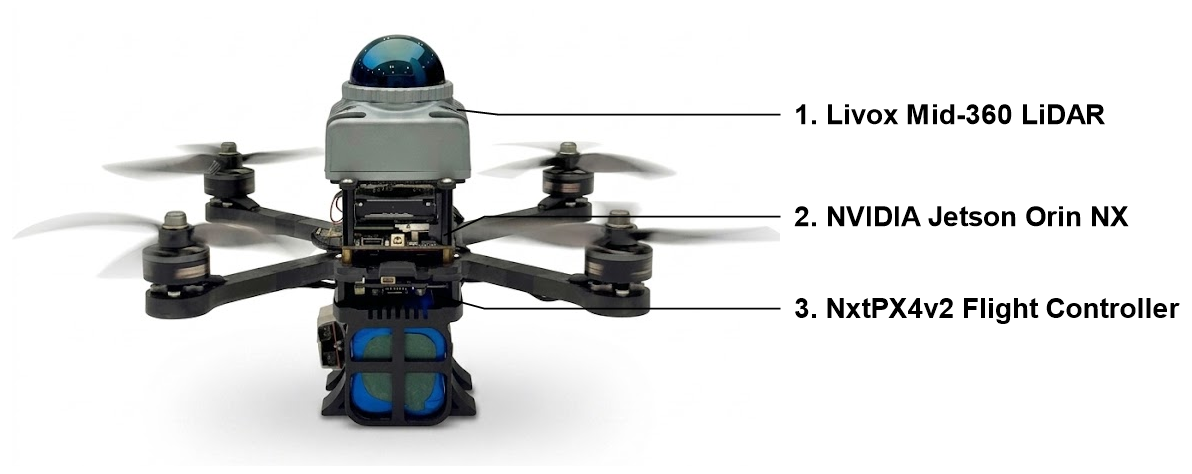}
    \caption{Quadrotor platform: (1) Livox Mid-360 LiDAR, (2) NVIDIA Jetson
    Orin NX, (3) NxtPX4v2 flight controller.}
    \label{fig:hardware_platform}
\end{figure}

\begin{figure*}[!t]
    \centering
    \includegraphics[width=\textwidth]{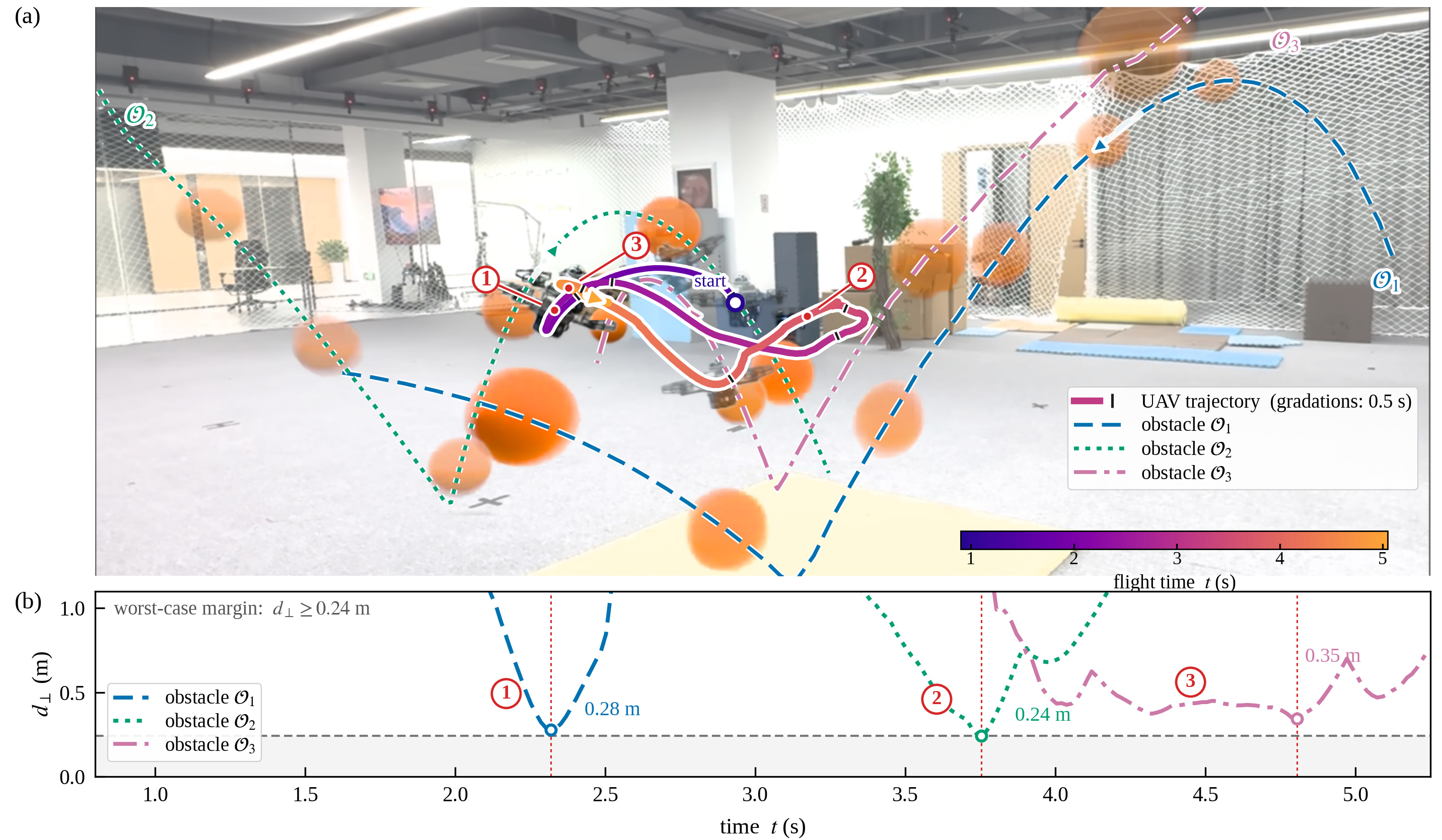}
    \caption{Real-world multi-threat trial in the cluttered indoor
    flight hall. (a) Three obstacles $\mathcal{O}_1$--$\mathcal{O}_3$ thrown in
    sequence; the executed quadrotor trajectory is colored by flight time, with
    closest-approach events 1--3 marked. (b) Surface-to-surface
    clearance $d_\perp(t)$ to each obstacle over time (accounting for platform
    and obstacle radii); the vehicle maintains a safe positive margin
    throughout, reaching a worst-case clearance of $0.24$~m.}
    \label{fig:dodging}
\end{figure*}

\subsection{Protocol and Baselines}
\label{sec:exp_protocol}

Each trial requires the quadrotor to navigate from a fixed start to a goal
through static clutter and dynamic projectiles.
A trial is counted as successful only if the vehicle reaches the goal collision-free
within the allotted time; physical contact and kinodynamic infeasibility
are recorded as failures.
Given our 20~Hz replanning loop, we define a deadline miss as any cycle in which planning
latency exceeds the timing budget $\Delta T_{\mathrm{budget}} = 50~\text{ms}$, forcing the
low-level controller to continue tracking the previously committed trajectory;
misses are counted per cycle and do not by themselves fail a trial.
Planning latency is measured from planning-input arrival to
trajectory commitment, covering tensor rasterization and network inference;
upstream sensing, detection, and track confirmation are shared by all methods and excluded.
Closed-loop simulations run in real time, and all methods share $50$ paired
evaluation seeds per condition, disjoint from the training seeds;
key success rates (SR) are reported with Wilson 95\% confidence intervals (CIs).
The entire pipeline runs onboard an NVIDIA Jetson Orin NX under realistic
multi-process resource contention on the custom quadrotor shown in
Fig.~\ref{fig:hardware_platform}. We supply identical synchronized inputs to
the state-of-the-art baselines FAPP~\cite{lu2024fapp} and
SIMP~\cite{huang2025safe}.

\subsection{Input Representation Ablation}
\label{sec:exp_representation}

To evaluate whether spatial depth alone can resolve dynamic threats under sparse sensing,
we benchmark three input variants under identical architectures and training budgets:
\emph{Depth-Only} (mask and velocity zeroed),
\emph{Depth+Mask} (velocity channels omitted),
and our full planning input tensor (Table~\ref{tab:representation}).
While Depth-Only navigates static corridors reliably,
it experiences severe collisions during crossing encounters ($22.0\%$ SR)
and rear pursuits ($26.0\%$ SR),
yielding an average dynamic collision rate (DCR, the fraction of trials with a dynamic collision) of $47.3\%$.
Adding the dynamic mask partially improves success in head-on encounters
but remains largely ineffective against lateral threats ($30.0\%$ SR).
In contrast, our full tensor achieves success rates of at least $92.0\%$ across all geometries,
reducing the average DCR to $4.0\%$.

This gap arises because range data alone cannot reveal lateral motion.
Between sparse sweeps, range signatures from a crossing obstacle
closely resemble those of thin static clutter,
causing depth-only models to respond to the threat only after
the viable reaction window has already closed.
While the dynamic mask confirms cluster association,
it lacks metric kinematic cues.
Encoding Cartesian surface velocities into the tensor
makes lateral relative motion directly observable,
enabling timely evasive maneuvers perpendicular to the obstacle's path.
\begin{table}[!tb]
    \centering
    \footnotesize
    \caption{Input representation ablation ($n=50$ trials per cell)}
    \label{tab:representation}
    \setlength{\tabcolsep}{3pt}
    \begin{tabularx}{\columnwidth}{@{}l*{5}{C}@{}}
        \toprule
        Method & \shortstack{Static\\SR $\uparrow$} & \shortstack{Head-on\\SR $\uparrow$} & \shortstack{Crossing\\SR $\uparrow$} & \shortstack{Rear\\SR $\uparrow$} & \shortstack{Avg.\\DCR $\downarrow$}\\
        \midrule
        Depth-Only & 96.0\% & 62.0\% & 22.0\% & 26.0\% & 47.3\%\\
        Depth+Mask & 96.0\% & 88.0\% & 30.0\% & 34.0\% & 36.7\%\\
        \bestrow \textbf{Ours} & \textbf{98.0\%} & \textbf{96.0\%} & \textbf{92.0\%} & \textbf{94.0\%} & \textbf{4.0\%}\\
        \bottomrule
    \end{tabularx}
\end{table}

\subsection{Dynamic Risk Field Ablation}
\label{sec:exp_risk}
To isolate the constituent mechanisms of the asymmetric dynamic risk field (Sec.~\ref{sec:risk_field}),
we evaluate four ablations against simulated future-collision ground truth:
\emph{No Dynamic Supervision},
\emph{No Future Prediction},
\emph{Isotropic Field} ($\sigma_{\parallel} = \sigma_{\perp}$),
and \emph{Symmetric Field} (the causal gate $\vect{v}_{\mathrm{rel}}^\top \vect{n}_{o}$ removed, the forecast retained).
Approaching-versus-receding (A--R) accuracy (Table~\ref{tab:risk}) compares the maximum candidate risk
between velocity-reversed frame pairs, labeled by the sign of the current Euclidean range rate.
Omitting the directional projection degrades A--R accuracy to $51.8\%$
and inflates the DCR to $34.0\%$;
we attribute this to residual penalties behind obstacles that have already
passed, which prolong evasive maneuvering against cleared obstacles.
Similarly, omitting constant-velocity horizon prediction severely degrades risk discrimination,
reducing the area under the receiver operating characteristic curve (AUC) for future-collision prediction to $0.735$.
In contrast, our full formulation combines along-track elongation of the risk field
with monotonic risk decay for receding obstacles,
achieving $96.2\%$ directional accuracy and reducing the DCR to $4.0\%$.

\begin{table}[!tb]
    \centering
    \caption{Dynamic risk field ablation (Avg.\ DCR as in Table~\ref{tab:representation})}
    \label{tab:risk}
    \footnotesize
    \setlength{\tabcolsep}{3.0pt}
    \begin{tabularx}{\columnwidth}{@{}l*{3}{C}@{}}
        \toprule
        Variant & \shortstack{A--R\\Acc. $\uparrow$} & \shortstack{Collision\\AUC $\uparrow$} & \shortstack{Avg.\\DCR $\downarrow$}\\
        \midrule
        No Dynamic Supervision & 58.4\% & 0.612 & 42.0\%\\
        No Future Prediction & 71.2\% & 0.735 & 28.0\%\\
        Isotropic Field & 86.5\% & 0.814 & 18.0\%\\
        Symmetric Field & 51.8\% & 0.648 & 34.0\%\\
        \bestrow \textbf{Ours} & \textbf{96.2\%} & \textbf{0.946} & \textbf{4.0\%}\\
        \bottomrule
    \end{tabularx}
\end{table}

\subsection{Computational Scalability}
\label{sec:exp_latency}
We benchmark computational scalability on the NVIDIA Jetson Orin NX
with $|\mathcal{O}_{t}| \in \{1, 4, 6\}$ simultaneous dynamic obstacles
(Table~\ref{tab:planning_efficiency}).
Benefiting from its lightweight analytical formulation,
FAPP achieves the lowest planning latency ($\sim 17$~ms at the 95th percentile, P95) with zero deadline misses.
In comparison, our dual-branch circular convolutional network and tensor processing incur a higher fixed computational overhead
under onboard multi-process contention,
exhibiting P95 latencies between $42.2$~ms and $45.8$~ms
and a $3.0\%$ deadline-miss rate at $|\mathcal{O}_{t}|=6$.
These misses are marginal: the worst case ($50.9$~ms) exceeds the budget by only $0.9$~ms, while the controller keeps tracking the previously committed trajectory.
Network inference remains nearly constant ($20.0$--$20.5$~ms P95);
the residual variation comes from rasterization, not inference.
Nonetheless, while SIMP suffers severe combinatorial explosion
with latencies exceeding $1400$~ms ($96.3\%$ misses),
our P95 latency stays within $45.8$~ms.
Although slower than FAPP, our planner spends this latency
on scoring every candidate primitive,
whose safety benefit is evaluated in Sec.~\ref{sec:exp_closed_loop}.
\begin{table}[!t]
    \centering
    \caption{Planning latency scalability}
    \label{tab:planning_efficiency}
    \footnotesize
    \setlength{\tabcolsep}{3pt}
    \begin{tabular*}{\columnwidth}{@{\extracolsep{\fill}}lccccc@{}}
        \toprule
         & \multicolumn{3}{c}{P95 latency (ms) $\downarrow$} & \multicolumn{2}{c}{$|\mathcal{O}|{=}6$}\\
        \cmidrule(lr){2-4}\cmidrule(l){5-6}
        Method & $|\mathcal{O}|{=}1$ & $|\mathcal{O}|{=}4$ & $|\mathcal{O}|{=}6$ & Max (ms) $\downarrow$ & Miss (\%) $\downarrow$\\
        \midrule
        SIMP & 1445.8 & 1861.5 & 2110.9 & 3237.5 & 96.3\\
        FAPP & \textbf{18.3} & \textbf{16.6} & \textbf{17.6} & \textbf{22.3} & \textbf{0.0}\\
        Ours & 45.8 & 42.2 & 45.6 & 50.9 & 3.0\\
        \bottomrule
    \end{tabular*}
\end{table}

\subsection{Multi-Threat Closed-Loop Safety}
\label{sec:exp_closed_loop}

We evaluate closed-loop flight safety in simulation across obstacle counts
$|\mathcal{O}| \in \{1, 4, 6\}$
and speeds $\|\vect{v}_o\| \in \{2.0, 6.0, 10.0\}$~m/s (Table~\ref{tab:closed_loop}).
While numerical baselines perform reliably in sparse, low-speed regimes,
their safety degrades sharply as scene density and encounter speeds increase.
Under the most adversarial condition with
$|\mathcal{O}|=6$ and $\|\vect{v}_o\| = 10$~m/s (Table~\ref{tab:closed_loop_hard}),
FAPP's mean clearance falls to $0.12$~m and its DCR reaches $86.0\%$,
although it incurs no deadline misses in the latency benchmark (Table~\ref{tab:planning_efficiency}).
We attribute this to local refinement of a single trajectory,
which limits the available evasive alternatives when multiple fast threats converge.
In contrast, by scoring every candidate primitive with offline-distilled risk,
our framework maintains a $0.46$~m mean clearance and an $84.0\%$ SR
(CI $[71.5, 91.7]$ vs.\ $[3.2, 18.8]$ for FAPP).

\begin{table*}[!t]
    \centering
    \footnotesize
    \caption{Closed-loop success rate (\%) under obstacle count $|\mathcal{O}|$ and speed $\|\vect{v}_o\|$ in m/s, $n=50$ trials per cell}
    \label{tab:closed_loop}
    \setlength{\tabcolsep}{5pt}
    \begin{tabularx}{\textwidth}{@{}l*{9}{C}@{}}
        \toprule
         & \multicolumn{3}{c}{$|\mathcal{O}|{=}1$} & \multicolumn{3}{c}{$|\mathcal{O}|{=}4$} & \multicolumn{3}{c}{$|\mathcal{O}|{=}6$}\\
        \cmidrule(lr){2-4}\cmidrule(lr){5-7}\cmidrule(l){8-10}
        Method & 2 & 6 & 10 & 2 & 6 & 10 & 2 & 6 & 10\\
        \midrule
        SIMP$^\dagger$ & 98.0 & 0.0 & 0.0 & 82.0 & 0.0 & 0.0 & 66.0 & 0.0 & 0.0\\
        FAPP & 100.0 & 90.0 & 74.0 & 88.0 & 62.0 & 26.0 & 74.0 & 40.0 & 8.0\\
        \bestrow \textbf{Ours} & \textbf{100.0} & \textbf{98.0} & \textbf{96.0} & \textbf{98.0} & \textbf{94.0} & \textbf{88.0} & \textbf{96.0} & \textbf{90.0} & \textbf{84.0}\\
        \bottomrule
    \end{tabularx}
    \vspace{1mm}
    \begin{minipage}{\linewidth}
        \footnotesize
        \(^\dagger\)SIMP failed all trials when \(\|\vect{v}_o\| \in \{6.0, 10.0\}\)~m/s due to optimization infeasibility at high closing speeds.
    \end{minipage}
\end{table*}

\begin{table}[!t]
    \centering
    \caption{Adversarial-regime avoidance ($|\mathcal{O}|{=}6$, $\|\vect{v}_o\|{=}10$~m/s, $n=50$)}
    \label{tab:closed_loop_hard}
    \footnotesize
    \setlength{\tabcolsep}{3pt}
    \begin{tabularx}{\columnwidth}{@{}l*{3}{C}c@{}}
        \toprule
        Method & SR $\uparrow$ & DCR $\downarrow$ & Other $\downarrow$ & Mean clearance $\uparrow$\\
        \midrule
        FAPP & 8.0\% & 86.0\% & 6.0\% & 0.12~m\\
        \bestrow \textbf{Ours} & \textbf{84.0\%} & \textbf{12.0\%} & \textbf{4.0\%} & \textbf{0.46~m}\\
        \bottomrule
    \end{tabularx}
    \vspace{0.6mm}
    \raggedright\footnotesize Other: remaining failures (static contact, infeasibility, or timeout).\par
\end{table}

\subsection{Real-World Flight Performance}
\label{sec:exp_hardware}

\begin{table}[!t]
    \centering
    \footnotesize
    \caption{Zero-shot real-world performance ($n=50$ trials per cell)}
    \label{tab:real_results}
    \setlength{\tabcolsep}{5pt}
    \begin{tabularx}{\columnwidth}{@{}l*{2}{C}@{}}
        \toprule
        Metric & Indoor & Outdoor\\
        \midrule
        Tennis SR $\uparrow$ & 94.0\% & 90.0\%\\
        Foam Ball SR $\uparrow$ & 86.0\% & 82.0\%\\
        Plush SR $\uparrow$ & 92.0\% & 88.0\%\\
        \midrule
        \bestrow \textbf{All-types SR $\uparrow$} & \textbf{90.7\%} & \textbf{86.7\%}\\
        Mean clearance $\uparrow$ & 0.42~m & 0.38~m\\
        \bottomrule
    \end{tabularx}
\end{table}

We deploy the trained policy with frozen weights
across two novel environments---a cluttered indoor flight hall and an outdoor field---against
unseen physical projectiles (a tennis ball, a foam basketball, and a plush toy)
at relative encounter speeds up to $15$~m/s (Table~\ref{tab:real_results}).
Fig.~\ref{fig:dodging} demonstrates continuous evasive maneuvers
against three sequentially thrown dynamic obstacles.
The weakly reflective foam basketball poses the greatest challenge,
inducing delayed track confirmation and tracking dropouts,
and yields the lowest success rates ($86.0\%$ indoor, $82.0\%$ outdoor).
We attribute the successful trials against it to the spatial margins embedded in the dynamic risk field,
which absorb perception uncertainty, and to simulation mask corruption
(Sec.~\ref{sec:di_corruption}), which reduces reliance on the binary mask
under tracking dropouts.
In the trial of Fig.~\ref{fig:dodging}, the quadrotor sustained a minimum clearance of
$0.24$~m, indicating that the policy transfers zero-shot by relying on metric relative motion.

\FloatBarrier
\section{Conclusion}

This paper presented a trajectory-risk learning framework
that enables agile quadrotors to evade fast-moving dynamic obstacles
using purely onboard sensing and computation.
By consolidating metric range panoramas, dynamic masks,
and Cartesian surface velocities into a unified planning input tensor,
the framework evaluates relative obstacle motion
via an obstacle-velocity-aligned asymmetric dynamic risk field.
Shifting risk evaluation and terminal-state optimization offline
into a dual-branch circular convolutional network reduces online planning
to a single forward pass over an omnidirectional anchor lattice.
The resulting P95 planning latency remains below the $50$~ms control budget
across obstacle counts, and scoring every candidate primitive sustains
avoidance success in dense, high-speed encounters.
Physical flight experiments confirmed zero-shot transfer with frozen weights,
demonstrating continuous evasion against unseen projectiles
at relative speeds up to $15$~m/s.
Extending this offline risk distillation paradigm
to decentralized multi-agent cooperative avoidance represents a promising future direction.

\balance
\bibliographystyle{IEEEtran}
\bibliography{references}

\end{document}